\documentclass{article}

\PassOptionsToPackage{numbers, compress}{natbib}
\usepackage[preprint]{main}

\usepackage[utf8]{inputenc} % allow utf-8 input
\usepackage[T1]{fontenc}    % use 8-bit T1 fonts
\usepackage{hyperref}       % hyperlinks
\usepackage{url}            % simple URL typesetting
\usepackage{booktabs}       % professional-quality tables
\usepackage{amsfonts}       % blackboard math symbols
\usepackage{nicefrac}       % compact symbols for 1/2, etc.
\usepackage{microtype}      % microtypography
\usepackage{graphicx}
\usepackage{algorithm}
\usepackage{amsmath}
\usepackage{tabularx}
\usepackage{subcaption}
\usepackage{colortbl}
\usepackage[table]{xcolor}
\newfloat{figtab}{htb}{fgtb}
\makeatletter
  \newcommand\figcaption{\def\@captype{figure}\caption}
  \newcommand\tabcaption{\def\@captype{table}\caption}
\makeatother
\usepackage{multirow}
\usepackage{wrapfig}
\usepackage{amssymb}% http://ctan.org/pkg/amssymb
\usepackage{pifont}% http://ctan.org/pkg/pifont
\usepackage{threeparttable}

\usepackage{xcolor}         % colors
 \usepackage{amsmath}
\usepackage{colortbl}
\definecolor{mygray}{rgb}{0.941, 0.941, 0.956}

\def\logo{\makebox[24pt][l]{\raisebox{-0.9ex}{\includegraphics[height=26pt]{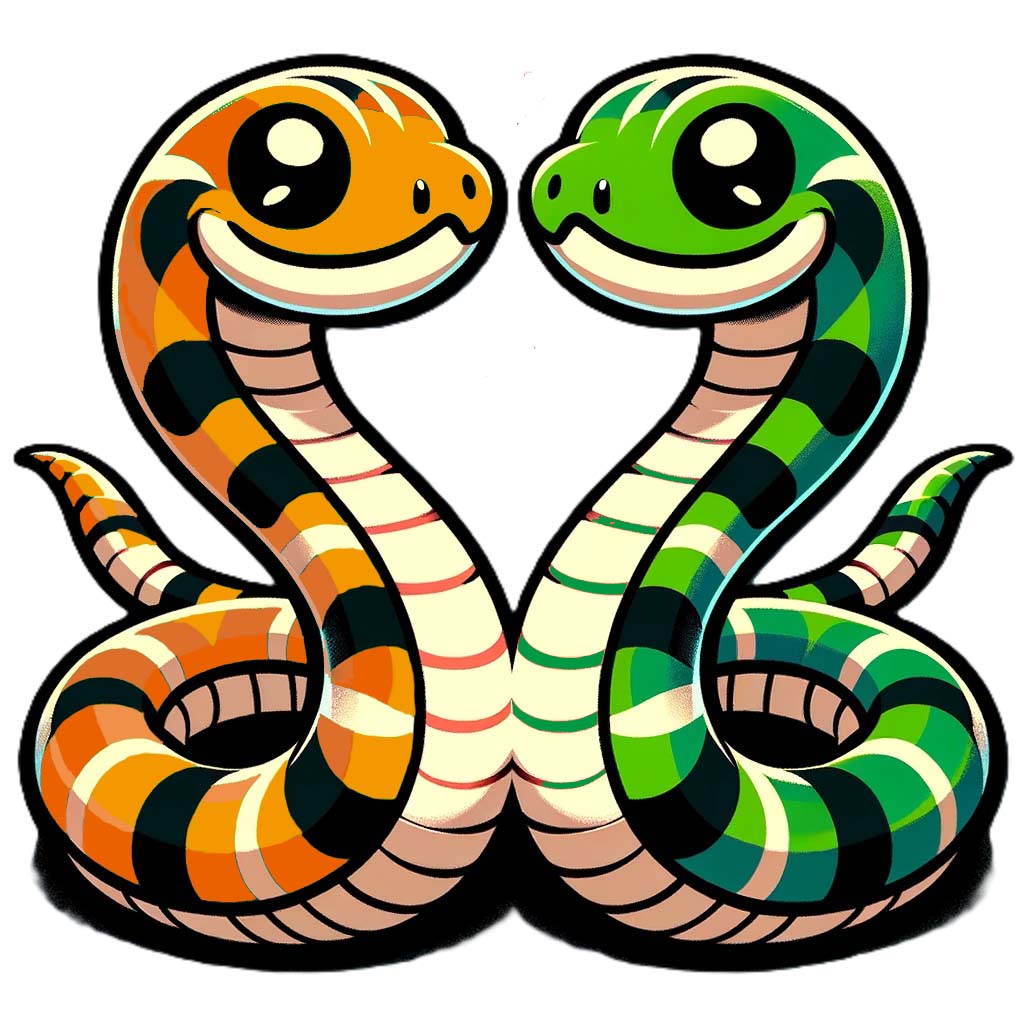}}}\hspace{5pt}}

\def\ie{\emph{i.e. }} 

\def\etc{\emph{etc. }}

\title{\logo  VL-Mamba: Exploring State Space Models for Multimodal Learning}

\author{%   
Yanyuan Qiao$^{1}$, Zheng Yu$^{1}$, Longteng Guo$^{2}$, Sihan Chen$^{2,3}$ \\[0pt]
\textbf{{Zijia Zhao}$^{2,3}$, {Mingzhen Sun}$^{2,3}$ , {Qi Wu} $^{1}\thanks{Corresponding author: Qi Wu}$, {Jing Liu}$^{2,3}$} \\[0pt]
$^1$Australian Institute for Machine Learning, The University of Adelaide\\[0pt]
$^2$Institute of Automation, Chinese Academy of Sciences\\[0pt]
$^3$School of Artificial Intelligence, University of Chinese Academy of Sciences \\[0pt]
{\tt\small \{yanyuan.qiao, zheng.yu, qi.wu01\}@adelaide.edu.au} \\[0pt]
{\tt\small \{longteng.guo, sihan.chen, jliu\}@nlpr.ia.ac.cn 
\{zijia.zhao, mingzhen.sun\}@ia.ac.cn}\\[5pt]
Project URL:  
{\small \url{https://yanyuanqiao.github.io/vl-mamba}}
}

\begin{document}

\maketitle

\begin{abstract}
Multimodal large language models (MLLMs) have attracted widespread interest and have rich applications. However, the inherent attention mechanism in its Transformer structure requires quadratic complexity and results in expensive computational overhead. Therefore, in this work, we propose VL-Mamba, a multimodal large language model based on state space models, which have been shown to have great potential for long-sequence modeling with fast inference and linear scaling in sequence length. Specifically, we first replace the transformer-based backbone language model such as LLama or Vicuna with the pre-trained Mamba language model. Then, we empirically explore how to effectively apply the 2D vision selective scan mechanism for multimodal learning and the combinations of different vision encoders and variants of pretrained Mamba language models. The extensive experiments on diverse multimodal benchmarks with competitive performance show the effectiveness of our proposed VL-Mamba and demonstrate the great potential of applying state space models for multimodal learning tasks. 
\end{abstract}

\section{Introduction}
\label{sec:intro}
Multimodal large language models (MLLM) have received widespread attention from the research community in recent years. 
It inherits the advanced capabilities of Large Language Models (LLMs) such as powerful language expression and logical reasoning. 
The integration of visual and textual information not only enhances the understanding of visual content but also provides a more comprehensive context for language understanding and generation. MLLM has shown great potential in solving visual problems in the real world and has rich applications in the fields of vision and language, such as image captioning~\cite{Karpathy2014DeepVA,Vinyals2014ShowAT}, referring expression comprehension (REC)~\cite{yu2018mattnet,qiao2020referring}, visual question answering (VQA)~\cite{Agrawal2015VQAVQ,Schwenk2022AOKVQAAB}, etc. Leveraging Transformer-based architectures~\cite{Vaswani2017AttentionIA} and large amounts of training data from web sources, MLLM has become a fundamental component in artificial intelligence research.

Although Transformers improve the ability of long-range dependencies and greatly enhance the performance of the model, this architecture is usually very computationally intensive. This is due to the inherent computational and memory complexity of the self-attention mechanism used by Transformer. The computational burden and memory requirements increase quadratically with the sequence length. 

To solve the bottleneck of long sequence modeling, the state space model (SSM) has been widely studied~\cite{LSSL, s5}.
It can be seen as a blend of recurrent neural networks (RNNs) and convolutional neural networks (CNNs).
Among these studies, the representative works are structured state space (S4)~\cite{s4} and its variants~\cite{s5, gupta2022diagonal-dss, S4D}.
The latest work Mamba~\cite{gu2023mamba} further improves S4, with a selection mechanism that allows the model to select relevant information in an input-dependent manner, combined with a hardware-aware implementation to achieve efficient training and inference.
Mamba outperforms Transformer on large-scale data and enjoys linear scaling in sequence length, which has proven to be a promising alternative to Transformer for language modeling.
Some concurrent works extended this architecture from 1D language to 2D vision domain~\cite{Ma2024UMambaEL,Liu2024VMambaVS,Yang2024VivimAV} such as image classification, biomedical image segmentation, \etc 
To the best of our knowledge, no work has explored how to utilize this efficient architecture to solve multimodal tasks. 

Inspired by the successes of SSM, in the paper, we introduce VL-Mamba, the first work that utilizes state space models for multimodal learning tasks. 
To be specific, as illustrated in Fig.~\ref{fig:vl-mamba}, we leverage the pre-trained Mamba language model as our backbone language model instead of conventional Transformer-based language models such as LLama~\cite{Touvron2023LLaMAOA} or Vicuna~\cite{vicuna2023}. 
Furthermore, we empirically explore the way to apply 2D vision selective scan mechanisms for VL-Mamba and introduce a novel MultiModal Connector (MMC) architecture, comprising a Vision Selective Scan (VSS) module and two linear layers, tailored to enrich the 2D-causal modeling of visual sequences. 
For the VSS module, we explore two distinct scan mechanisms: the Bidirectional-Scan Mechanism (BSM) and the Cross-Scan Mechanism (CSM). The BSM conducts scans of visual sequences in both forward and backward directions, while the CSM extends scanning capability to four directions.
In addition, we study the combinations of different vision encoders, variants of pretrained Mambe language models, and multimodal connectors to find the effect of different components for VL-Mamba.
Extensive experiments are conducted on various multimodal learning benchmarks to verify the effectiveness of VL-Mamba. Our model achieves competitive performance with other small MLLMs of similar size and even outperforms large MLLMs (e.g., 7B and 13B versions of LLaVA-1.5~\cite{liu2023improvedllava}) on some popular benchmarks.

In summary, our contributions are as follows:
\begin{itemize}
    \item We propose VL-Mamba, which is the first work to explore and exploit the state space model in solving multimodal learning tasks, which provides a novel framework option for multimodal large language models other than transformer-based architectures.
    \item We empirically explore the effect of different components for VL-Mamba and introduce a novel MultiModal Connector containing a Vision Selective Scan (VSS) module to improve the representational capabilities.
    \item We conduct extensive experiments on diverse multimodal learning benchmarks. The experiments demonstrate that VL-Mamba achieves competitive performance compared to existing multimodal large language models.
    \item We make the code open source to promote the research of applying state space models for multimodal learning.
\end{itemize}

\section{{Related Work}}
\label{sec:related work}
\subsection{State Space Models (SSMs)}
Modern state space models (SSMs) are derived from the classical state space model~\cite{kalman1960new} and have become an efficient building block for constructing deep networks, thereby achieving cutting-edge performance in analyzing continuous long-sequence data. They particularly excel at capturing long-range dependencies (LRDs) and leveraging parallel training methods to increase efficiency.
Initiated by a HiPPO matrix~\cite{gu2020hippo}, Linear State Space Layer (LSSL)~\cite{LSSL} combines the advantages of continuous-time models (CTMs), RNNs, and CNNs, which demonstrates the potential of deep SSMs to solve long-range dependencies. However, the practical feasibility of LSSL is hampered by the large computational and memory requirements imposed by the state representation.
Subsequently, the Structured State Space (S4)~\cite{s4} addresses the main computational bottleneck in prior research. This is achieved through novel parameterizations catering to continuous-time, recurrent, and convolutional views of the state space model, thereby effectively modeling long-range dependencies.
S4 has subsequently seen some variants~\cite{s5, gupta2022diagonal-dss, S4D}, such as the Diagonal State Space (DSS) model~\cite{gupta2022diagonal-dss}, which forces the state matrix to be a diagonal matrix, making it easier to formulate, implement, and analyze, and can be proven to be as expressive as a general state space, while S4D~\cite{S4D} provides a new mathematical analysis for DSS initialization, making it simpler and more efficient.

A recent work, named Mamba~\cite{gu2023mamba}, further improves S4 with a selection mechanism that incorporates time-varying parameters into SSM, allowing the model to select relevant information in an input-dependent manner. It proposes a hardware-aware algorithm to achieve efficient training and inference. Mamba's superior scaling performance shows that it is a promising alternative to the Transformer in long-sequence modeling. Many works extend Mamba from Natural Language Processing (NLP) to other fields~\cite{Yang2024VivimAV, Xing2024SegMambaLS,ruan2024vm}.
Vision Mamba (Vim)~\cite{Zhu2024VisionME} applies Mamba to the Vision Transfomer (ViT) architecture, and combines bidirectional SSM for data-dependent global visual context modeling and position embedding for location-aware visual understanding.
Visual State Space Model (VMamba)~\cite{Liu2024VMambaVS} designs a cross-scan mechanism to bridge the gap between 1-D array scanning and 2-D plain traversing.
U-Mamba~\cite{Ma2024UMambaEL} proposes a hybrid CNN-SSM architecture to capture both localized fine-grained features and long-range dependencies in images, to solve the biomedical image segmentation task.
{In this work, we explore how to transfer the success of Mamba to solve the more challenging multimodal learning tasks, which often require modeling of both vision and language modalities and complex reasoning.}

\subsection{Multimodal Large Language Model (MLLM)}
With the development of the powerful Large Language Models (LLMs)~\cite{Touvron2023LLaMAOA,Zhang2022OPTOP,Chowdhery2022PaLMSL}, 
many studies~\cite{achiam2023gpt4,Driess2023PaLMEAE,chen2023minigptv2,Qwen-VL,ye2023mplug,Chu2023MobileVLMA} extend LLMs to multimodal domains by combining visual input with LLM to build the multimodal large language model (MLLM).
Flamingo~\cite{alayrac2022flamingo} freezes pre-trained visual encoders and large language models and fuses visual and language modalities with gated cross-attention, demonstrating excellent few-shot learning performance.
BLIP~\cite{Li2022BLIPBL} uses a dataset bootstrapped from large-scale noisy image-text pairs to pre-train a multi-modal mixture of encoder-decoder models by injecting different synthetic captions and removing noisy captions.
Based on this, BLIP-2~\cite{Li2023BLIP2BL} uses Querying Transformer (Q-Former) to bridge the modal gap.
InstructBLIP~\cite{instructblip} further proposes an instruction-aware visual feature extraction mechanism that can flexibly and effectively extract visual information features according to the given instructions.
LLaVA~\cite{liu2023improvedllava, liu2023llava} leverages advanced LLMs (\ie LLaMA~\cite{Touvron2023LLaMAOA} and Vicuna~\cite{vicuna2023}) as the language model and CLIP~\cite{Radford2021LearningTV} as the visual encoder, it transforms visual tokens into language tokens with a simple MLP layer.
MiniGPT-4~\cite{zhu2023minigpt} directly aligns visual information with the language model to accomplish diverse vision-language tasks without using external vision models. Usually, the training of MLLMs contains two stages, of which the first stage is to pretrain the model on a large collection of image-text pairs to acquire the alignment of vision-language knowledge, and the second stage is to finetune the model with a smaller but high-quality multimodal instruction tuning dataset with a designed conversational template.

These MLLM works have greatly advanced research in the fields of computer vision and natural language processing. However, since the main framework of these models relies on Transformers, the attention mechanism in Transformers inherently has high computational complexity in inference for long sequences. 
To alleviate the abovementioned issues related to modeling long-range sequences in the area of multi-modal learning, we propose the VL-Mamba, which is based on the state space model. To be specific, we utilize pretrained Mamba~\cite{gu2023mamba} language model as our backbone language model, rather than Transformer-based LLMs such as LLama~\cite{Touvron2023LLaMAOA} or Vicuna~\cite{vicuna2023}. Moreover, we empirically explore the effective application of 2D selective scan mechanism in the multimodal VL-Mamba and the combination of different vision encoders and variants of Mamba language models.

\section{Method}
\label{sec:method}
In this section, we first introduce the preliminary concepts of state space models (Sec. \ref{subsec:pre}). Then, we describe the details of our proposed VL-Mamba (Sec. \ref{subsec:model}), which mainly includes the Vision Encoder, MultiModal Connector, and the Mamba LLM.

\subsection{Preliminaries}
\label{subsec:pre}
State space models (SSMs) are commonly considered linear time-invariant systems that map stimulation $x(t) \in \mathbb{R}^L$ to response $y(t) \in \mathbb{R}^M$ through a hidden state $h(t) \in \mathbb{R}^N$. Mathematically, these models are typically formulated as linear ordinary differential equations (ODEs), where the parameters include $ \mathbf{A} \in \mathbb{C}^{N \times N}$, $ \mathbf{B} \in \mathbb{C}^{N}$ for a state size $N$, and the skip connection $ \mathbf{D} \in \mathbb{C}^1$. The system dynamics and output equations are given by:

\begin{equation}
\begin{aligned}
\label{eq:lti}
h'(t) &= \mathbf{A}h(t) + \mathbf{B}x(t), \\
y(t) &= \mathbf{C}h(t) +  \mathbf{D}h(t).
\end{aligned}
\end{equation}

Subsequently, the process of discretization is commonly employed to incorporate Eq. \ref{eq:lti} practical deep learning algorithms.
In this context, $\mathbf{\Delta}$ represents the timescale parameter that is used to convert the continuous parameters $\mathbf{A}, \mathbf{B}$ into discrete parameters, $\mathbf{\bar{A}}, \mathbf{\bar{B}}$. The zero-order hold (ZOH) method is commonly utilized for this discretization, and it is described as follows:
\begin{equation}
\begin{aligned}
\label{eq:zoh}
\mathbf{\overline{A}} &= \exp{(\mathbf{\Delta}\mathbf{A})}, \\
\mathbf{\overline{B}} &= (\mathbf{\Delta} \mathbf{A})^{-1}(\exp{(\mathbf{\Delta} \mathbf{A})} - \mathbf{I}) \cdot \mathbf{\Delta} \mathbf{B}.
\end{aligned}
\end{equation}

Once discretized, Eq.~\ref{eq:zoh} can be reformulated with the step size 
$\Delta$ as:
\begin{equation}
\begin{aligned}
\label{eq:discrete_lti}
h_t &= \mathbf{\overline{A}}h_{k-1} + \mathbf{\overline{B}}x_{k}, \\
y_t &= \mathbf{C}h_k + \mathbf{D}x_k.
\end{aligned}
\end{equation}

Nevertheless, the formulation in \ref{eq:discrete_lti} is predicated on a Linear Time Invariance (LTI) system where parameters are invariant despite changes in the input. To address this constraint, the recent work Mamba~\cite{gu2023mamba} explored integrating a selective scan technique, in which the matrices $\mathbf{\overline{B}}$, $ \mathbf{C}$, and $\mathbf{\Delta}$ are derived from the input data. This change equipped Mamba with the ability to dynamically focus on information from the input sequence, which increased the model's capability.

\begin{figure}[t]
  \centering
  \includegraphics[width=1\linewidth]{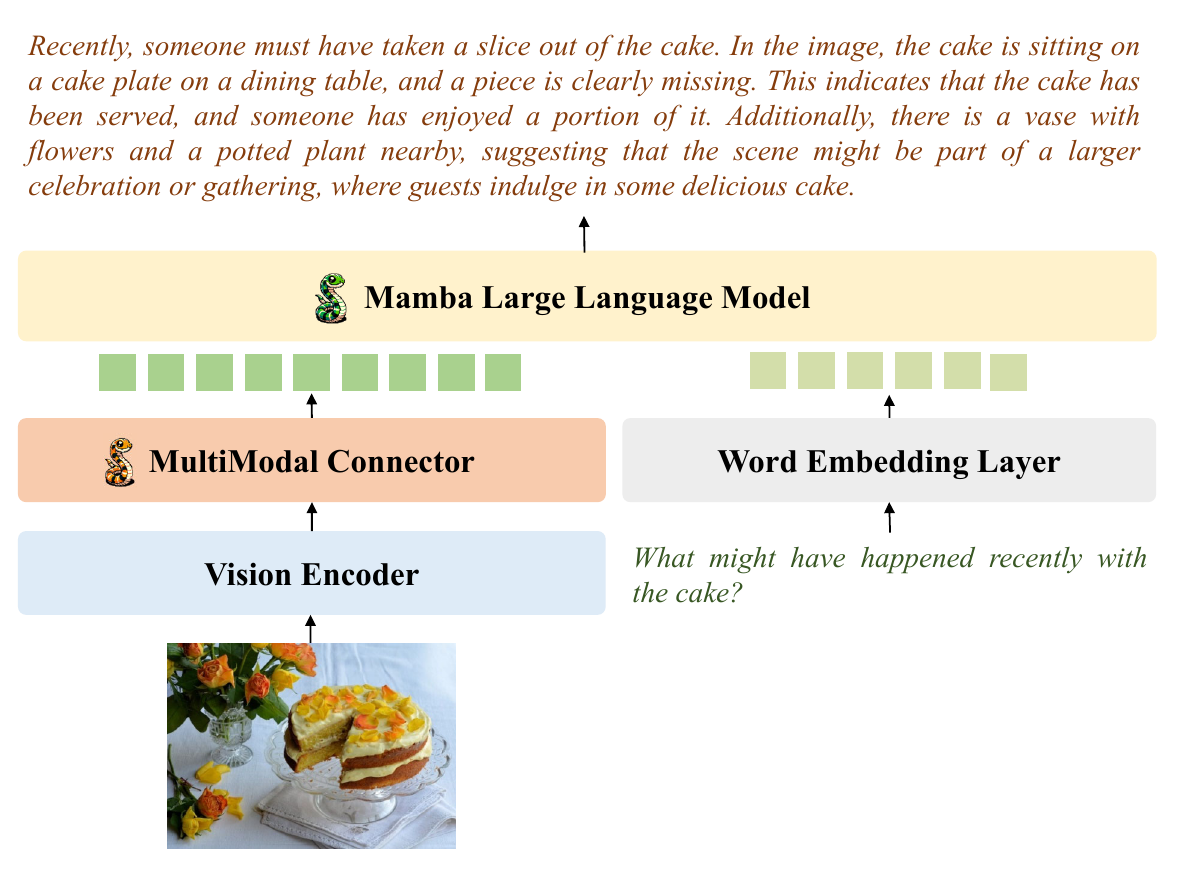}
  \caption{The architecture of VL-Mamba. It contains a Vision Encoder, a MultiModal Connector (MMC), and a language model. We utilize the pre-trained Mamba Large Language Model (Mamba LLM) as its language model, and the pre-trained Vision Transformer model as its vision encoder.
  }
  \label{fig:vl-mamba}
\end{figure}

\subsection{VL-Mamba Model}
\label{subsec:model}

\subsubsection{Overall Architecture}
\label{subsubsec:all}
The architecture of VL-Mamba consists of a pretrained vision encoder, a randomly initialized MultiModal Connector (MMC) which incorporates the 2D vision selective scan mechanism, and a pretrained Mamba Large Language Model (Mamba LLM), as illustrated in Fig.~\ref{fig:vl-mamba}.
Taking an image as input, we first obtain visual features through the visual encoder, then feed the visual sequences into MMC, and then this output vector combined with a tokenized text query is fed into Mamba LLM to generate the corresponding response.

\subsubsection{Vision Encoder}

The vision encoder of VL-Mamba uses the Vision Transformer (ViT)~\cite{vit} architecture that generates a sequence of patch features from raw images.
The vision encoder ${f_V}$, takes an image $I$ as input and produces a sequence of the visual patch features $V_{img}$, as follows: 

\begin{equation}
\begin{aligned}
\label{eq:vit}
V_{img} = {f_V}(I).
\end{aligned}
\end{equation}

\begin{figure}[t]
  \centering
  \includegraphics[width=0.99\linewidth]{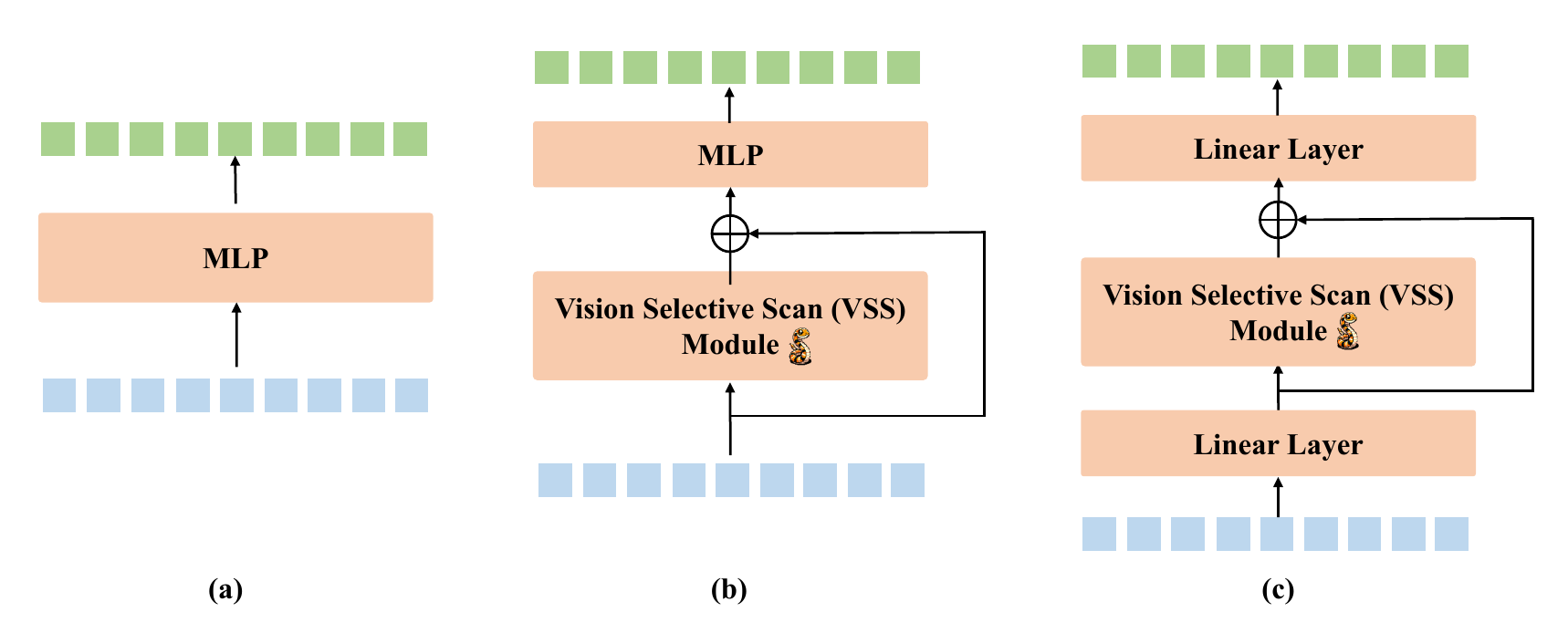}
  \caption{{Three architectures of MultiModal Connector: (a) MLP; (b) MLP-VSS; (c) VSS-2 Linear Layer. }
  }
  \label{fig:mmp}
  \vspace{-8pt}
\end{figure}

\begin{figure}[t]
 \vspace{-10pt}
  \centering
  \includegraphics[width=0.99\linewidth]{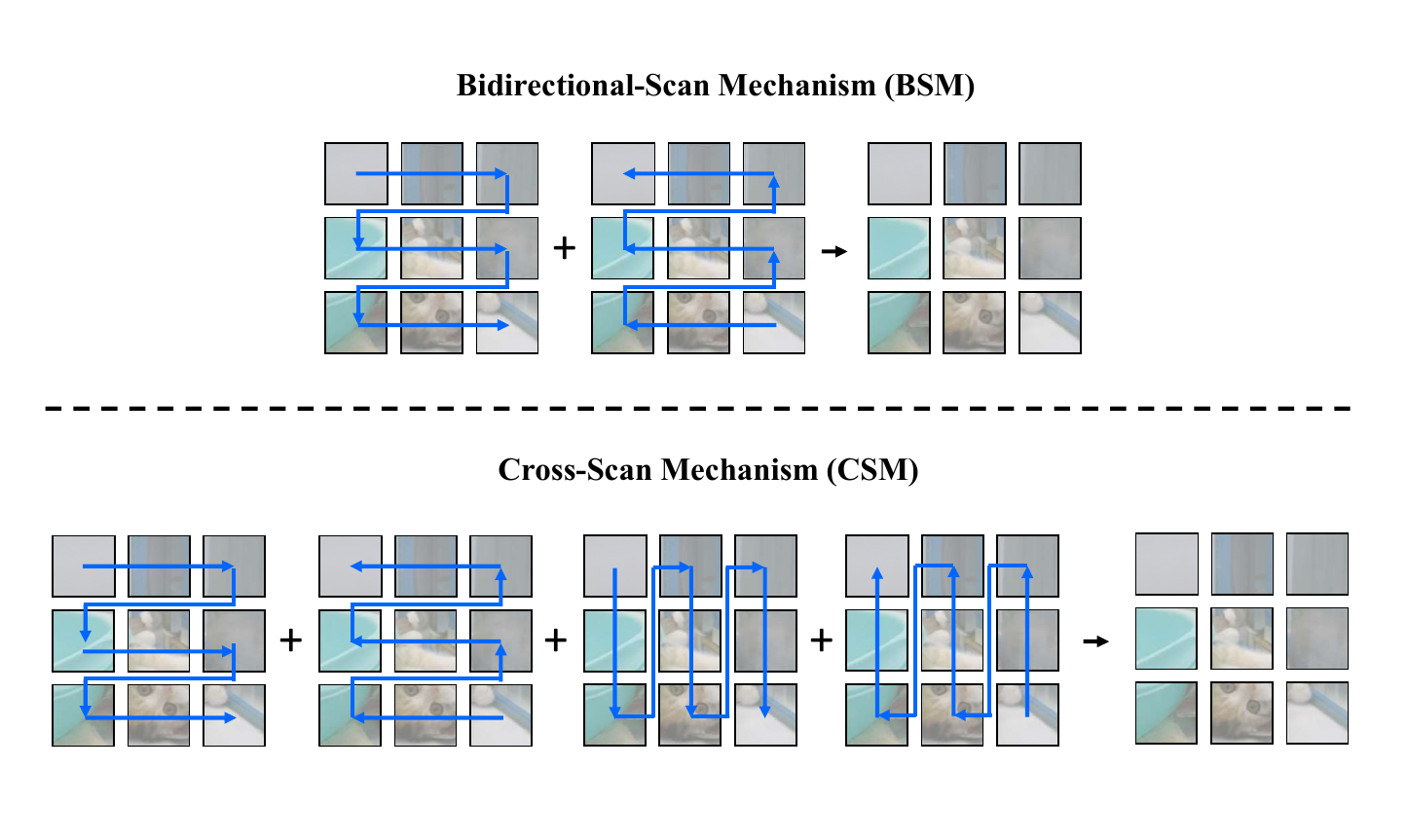}
  \caption{Illustration of two different Vision Selective Scan (VSS) Mechanisms: Bidirectional-Scan Mechanism (BSM) (top) and Cross-Scan Mechanism (CSM) (bottom).
  }
  \label{fig:2D scan}
\end{figure}

\subsubsection{{MultiModal Connector (MMC)}}

Since the state space models are designed to process 1D sequential data such as language sequences that have causal relationships, but the visual sequences generated by the vision encoder are non-causal data, 2D vision selective scan mechanisms are proposed to solve computer vision tasks. In this work, we try to apply the 2D vision selective scan mechanisms for multimodal learning by ensembling them in the multimodal connector of VL-Mamba.
Specifically, we explore three variants of multimodal connectors: 
\begin{itemize}
    \item \textbf{MLP}: a two-layer Multi-Layer Perceptron (MLP), which is depicted in Fig.~\ref{fig:mmp}(a).
    \item \textbf{VSS-MLP}: a Vision Selective Scan (VSS) module combined with an MLP. The architecture is shown in Fig.~\ref{fig:mmp}(b).
    \item \textbf{VSS-L2}: a VSS module combined with two linear layers, which is depicted in Fig.~\ref{fig:mmp}(c). 
\end{itemize}

The VSS module aims to bridge the gap between the 1D sequential processing capabilities inherent in the SSM and the 2D non-causal visual information. Specifically, the VSS module consists of a 2D vision scan mechanism and one mamba layer. 
In this work, we utilize two 2D scan mechanisms: Bidirectional-Scan Mechanism and Cross-Scan Mechanism, as follows: %, which is illustrated in~\Cref{fig:2D scan}.

\begin{itemize}
    \item \textbf{Bidirectional-Scan Mechanism (BSM)} scans the image patch features in both forward and backward directions, which aims to capture a broader context without increasing computational complexity, as illustrated in the top of Fig.~\ref{fig:2D scan}.
    \item \textbf{Cross-Scan Mechanism (CSM)} unfolds image patch features into sequences along rows and columns and scans them in four directions (diagonally across the image), as shown in the bottom of Fig.~\ref{fig:2D scan}. 
\end{itemize}
After the scan process, these sequences are passed through the mamba layer and reshaped back into the original image patch order, and all such features are merged to form a comprehensive representation.

As shown in Fig.~\ref{fig:mmp}(b), the input of the multimodal connector is the sequential image patch features $V_{img}$ extracted from the input images via the transformer-based vision encoder. These feature vectors are then passed through a Vision Selective Scan (VSS) module to obtain the visual scanned feature $V_{scan}$.
After the VSS module, the output vectors $V_{scan}$ are combined with the original image patch features $V_{img}$ through a skip connection. The combined vector is then passed into a norm layer and a two-layer Mult-Layer (MLP): 

\begin{equation}
\begin{aligned}
\label{eq:mmc}
V_{scan} &= \mathbf{VSS}(V_{img}), \\
V_{out} &= \mathbf{MLP}(\mathbf{Norm}(V_{scan} + V_{img})).
\end{aligned}
\end{equation}

And for the variant MMC in Fig.~\ref{fig:mmp}(c), the feed-forward pass progress can be formulated as follows:

\begin{equation}
\begin{aligned}
\label{eq:mmc}
V_{img}^{'} &= \mathbf{Linear}(V_{img}), \\
V_{scan} &= \mathbf{VSS}(\mathbf{GELU}(V_{img}^{'})), \\
V_{out} &= \mathbf{Linear}(\mathbf{Norm}(V_{scan} + V_{img}^{'})).
\end{aligned}
\end{equation}

\subsubsection{{Mamba LLM}}

We use the pre-trained Mamba Large Language Model (Mamba LLM)~\cite{gu2023mamba} ${f_{L}}$ as our language model. 
Given a natural language query $Q$, we utilize the tokenizer and embedding module $f_T$ to map the text input into the embedding space.
Then the visual vector $V_{out}$ and textual $T$ are concatenated and put into the MambaLLM to obtain the response $R$.

\begin{equation}
\begin{aligned}
\label{eq:llm}
R = {f_{L}}(V_{out}, f_T(Q)).
\end{aligned}
\end{equation}

\section{Experiment}
\label{sec:expri}
In this section, we first introduce our experimental setup including implementation details and MLLM benchmarks in Sec.~\ref{subsec:setup}.
Then we present the quantitative comparison and qualitative results in Sec.~\ref{subsec:sota} and Sec.~\ref{subsec:vis}. Finally, we conduct ablation studies in Sec.~\ref{subsec:abla}.

\subsection{Experimental Setup}
\label{subsec:setup}

\subsubsection{Implementation details}
Following~\cite{liu2023llava,liu2023improvedllava}, the training process contains two stages: vision-and-language alignment pre-training and multimodal instruction tuning.
During the pretraining stage, we freeze the vision encoder and Mamba LLM and only keep the multimodal connector updated.
Then we finetune both the multimodal connector and the Mamba LLM in the instruction tuning stage.
Our model is trained on 8 NVIDIA Tesla A800 GPUs.
%The training and hyperparameters setting details are illustrated in the Appendix.

\subsubsection{MLLM Benchmarks}
We evaluate our model on a diverse set of 8 benchmarks:
VQA-v2~\cite{goyal2017vqav2}, GQA~\cite{hudson2019gqa}, ScienceQA-IMG~\cite{lu2022learn}, TextVQA~\cite{singh2019textvqa}, POPE~\cite{li2023pope}, MME~\cite{fu2023mme}, MMBench~\cite{Liu2023MMBenchIY}, MM-Vet~\cite{yu2023mmvet}.
VQA-v2~\cite{goyal2017vqav2} evaluates models' ability to understand and reason about images and questions. 
GQA~\cite{hudson2019gqa} assesses spatial understanding and multi-step inference in real-world images. 
ScienceQA~\cite{lu2022learn} offers multimodal multiple-choice questions on scientific topics, requiring common sense reasoning. 
The questions in TextVQA~\cite{singh2019textvqa} are related to the text in an image, it evaluates the model's optical character recognition (OCR) and inference capabilities.
%TextVQA~\cite{singh2019textvqa} assesses models' OCR and reasoning capabilities by focusing on images with text. 
POPE~\cite{li2023pope} provides a benchmark for evaluating object hallucinations, which is a binary classification task that prompts the model to answer whether an object exists.
MME~\cite{fu2023mme} evaluates perceptual and cognitive abilities, including OCR, object recognition, common sense reasoning, numerical calculations, text translation, and code reasoning. 
MMBench~\cite{Liu2023MMBenchIY}  features 3,000 single-choice questions across 20 dimensions, using a CircularEval strategy for robust evaluation, with ChatGPT matching model predictions to choices. 
MM-Vet~\cite{yu2023mmvet} identifies 16 emergent tasks from core visual and linguistic (VL) capabilities, including Recognition, Knowledge, OCR, Spatial awareness, Language generation, and Math.

\begin{table}[!t]
\caption{\textbf{Comparison with SoTA methods on 8 benchmarks.} 
Benchmark names are abbreviated due to space limits. VQA-v2~\cite{goyal2017vqav2}; GQA~\cite{hudson2019gqa}; SQA$^\text{I}$: ScienceQA-IMG~\cite{lu2022learn}; VQA$^\text{T}$: TextVQA~\cite{singh2019textvqa}; POPE~\cite{li2023pope}; MME~\cite{fu2023mme}; MMB: MMBench~\cite{Liu2023MMBenchIY}; MM-Vet~\cite{yu2023mmvet}.
PT and IT indicate the number of samples in the pretraining and instruction tuning stages, respectively.
}
\label{tab:results}
\centering
\vspace{3pt}
\renewcommand{\arraystretch}{1.25}
%\scalebox{0.5}{
\resizebox{\linewidth}{!}{
\begin{tabular}{ll cc | cccc | cccc }
\toprule
Method & LLM  & PT & IT & VQA$^\text{v2}$ & GQA & SQA$^\text{I}$ & VQA$^\text{T}$ & POPE & MME & MMB & MM-Vet \\
\midrule
BLIP-2~\cite{Li2023BLIP2BL} & Vicuna-13B  & 129M & - & 41.0 & 41.0 & 61.0 & 42.5 & {85.3} & 1293.8 & -- & 22.4 \\
MiniGPT-4~\cite{zhu2023minigpt}&Vicuna-7B&5M&5K&-&32.2&-&-&-&581.7&23.0&-\\
InstructBLIP~\cite{instructblip} & Vicuna-7B & 129M & 1.2M & -- & 49.2 & 60.5 & 50.1 & -- & -- & 36 & 26.2 \\ 
InstructBLIP~\cite{instructblip} & Vicuna-13B  & 129M & 1.2M & -- & 49.5 & 63.1 & 50.7 & 78.9 & 1212.8 & -- & 25.6 \\ 
Shikra~\cite{chen2023shikra} & Vicuna-13B  & 600K & 5.5M & 77.4 & -- & -- & -- & -- & -- & 58.8 & -- \\ 
Otter~\cite{Li2023OtterAM} & LLaMA-7B & -& -&-&-&-&-&-& 1292.3 & 48.3 & 24.6\\
mPLUG-Owl~\cite{ye2023mplug} & LLaMA-7B  & 2.1M & 102K & -&-&-&-&-&967.3&49.4&-\\
IDEFICS-9B~\cite{laurenccon2024obelics} & LLaMA-7B & 353M & 1M & 50.9 & 38.4 & -- & 25.9 & -- & -- & 48.2 & -- \\ 
IDEFICS-80B~\cite{laurenccon2024obelics} & LLaMA-65B & 353M & 1M & 60.0 & 45.2 & -- & 30.9 & -- & -- & 54.5 & -- \\
Qwen-VL~\cite{bai2023qwen} & Qwen-7B  & 1.4B & 50M & 78.8 & 59.3 & 67.1 & 63.8 & -- & -- & 38.2 & -- \\ 
Qwen-VL-Chat~\cite{bai2023qwen}  & Qwen-7B & 1.4B & 50M & 78.2 & 57.5 & 68.2 & 61.5 & -- & 1487.5 & 60.6 & -- \\ 
LLaVA-1.5~\cite{liu2023llava} & Vicuna-7B & 558K&665K&78.5&62.0&66.8&58.2&85.9&1510.7&64.3&30.5\\
LLaVA-1.5~\cite{liu2023llava} & Vicuna-13B & 558K&665K&80.0&63.3&71.6&61.3&85.9&1531.3&67.7&35.4\\
\midrule
LLaVA-Phi~\cite{zhu2024llavaphi} & Phi-2-2.7B & 558K& 665K& 71.4 & - &{68.4} & 48.6 & 85.0 & 1335.1 & {59.8} & 28.9\\
MobileVLM-3B~\cite{Chu2023MobileVLMA} & MobileLLaMA-2.7B  & 558K & 665K& - & {59.0} & 61.2 &47.5 & 84.9 & 1288.9 & 59.6 & -\\
\midrule
\rowcolor{mygray}
VL-Mamba& Mamba LLM{-2.8B} &558K & 665K&{76.6} & 56.2 & 65.4 & {48.9} &{84.4} &{1369.6}& 57.0& {32.6}\\
\bottomrule
\end{tabular}}
\end{table}

\begin{figure}[h]
  \centering
  \includegraphics[width=0.99\linewidth]{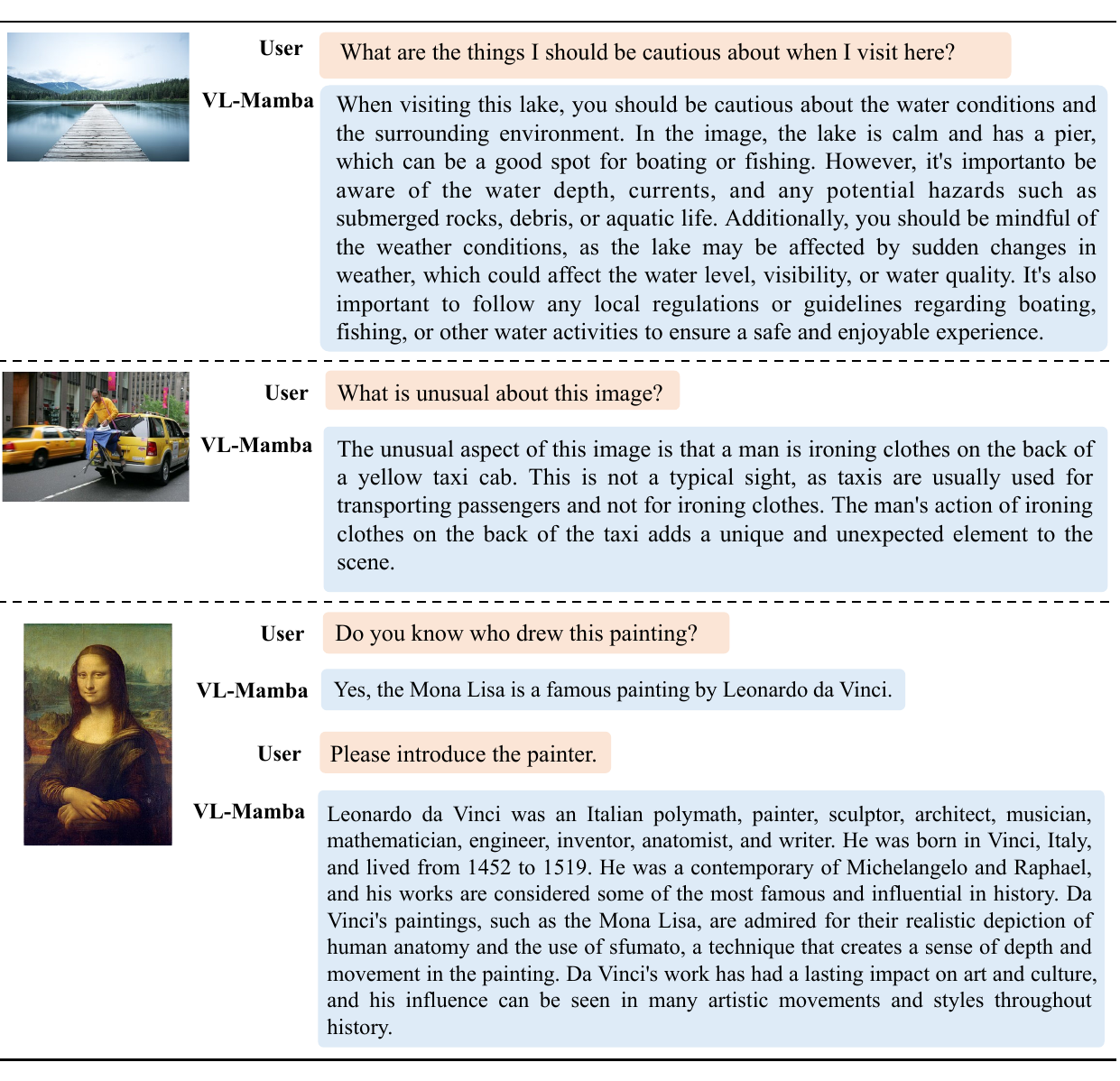}
  \caption{Examples of response generated by VL-Mamba.
  }
  \label{fig:vis}
\end{figure}

\subsection{Quantitative Evaluation}
\label{subsec:sota}
As is shown in Table~\ref{tab:results}, we compare our proposed model VL-Mamba with some SoTA multimodal large language models.
Compared with the MobileVLM-3B~\cite{Chu2023MobileVLMA} model with similar scale parameters and the same amount of multimodal training data, our model surpasses the performance on SQA$^\text{I}$ (65.4 v.s. 61.2), VQA$^\text{T}$ (48.9 v.s. 47.5), and MME (1369.6 v.s. 1288.9), though the Mamba LLM uses much less pretrained tokens (627B) than the backbone MobileLLaMA (1.3T) of MobileVLM. 
Compared with the LLaVA-Phi~\cite{zhu2024llavaphi} model with a SoTA language model Phi-2-2.7B with 1.4T pretrained tokens, our performance shows superior on VQA-v2 (76.6 v.s. 71.4), MME (1369 v.s. 1335.1), and MM-Vet (32.6 v.s. 28.9).
It is worth noting that though our proposed model has fewer parameters and limited training data, it also achieves comparable performance compared with some models with a larger number of parameters. Its performance on the POPE benchmark is similar to LLaVA-1.5~\cite{liu2023improvedllava}, where the LLM parameters are 13B, which is approximately 4.6 times larger than the Mamba LLM. 
These promising results demonstrate the effectiveness of our proposed VL-Mamba and show the potential of utilizing the state space models in multimodal learning tasks.

\subsection{Qualitative Result}
\label{subsec:vis}
We present some examples to see the qualitative results of the VL-Mamba. As shown in Fig.~\ref{fig:vis}, the VL-Mamba could well understand the user's question and respond accurately.

\subsection{Ablation Study}
\label{subsec:abla}

\subsubsection{The Effect of Variants of Language Model}
Table~\ref{tab:lang} shows the ablation experiment of evaluating the effectiveness of different variants of the language model.
We conduct experiments on three different variants, Mamba-1.4B which has 1.4B parameters and is trained on Pile~\cite{gao2020pile} with 300B tokens, Mamba-2.8B-Pile with 2.8B parameters and trained on Pile 300B tokens and Mamba-2.8B-Slimpj trained on SlimPajama with 627B tokens. Specifically, we construct the baseline models by using the same variant CLIP-ViT as the vision encoder, Mamba language models as backbone large language models, and vanilla MLP MultiModal Connectors without 2D vision selective scan modules. We can see with the increase of model scale and training tokens, Mamba-2.8B-Slimpj outperforms the other two variants on all benchmarks. Thus, we choose Mamba-2.8B-Slimpj for other experiments. 

\begin{table*}[th!]
\caption{Ablation study of the variants of the language model.
}
\label{tab:lang}
\centering
\renewcommand{\arraystretch}{1.15}
%\scalebox{0.76}{
\resizebox{0.9\linewidth}{!}{
\begin{tabular}{l| cccc | cccc }
\toprule
Method & VQA$^\text{v2}$ & GQA & SQA$^\text{I}$ & VQA$^\text{T}$ & POPE & MME & MMB & MM-Vet \\
\midrule
 Mamba-1.4B &71.7 & 49.9 & 56.1& 42.6 &84.5 & 1277.7& 46.9& 24.0\\
 Mamba-2.8B-Pile &73.6 & 53.0 & 60.8& 42.7 &84.7 & 1321.3& 52.1& 28.5\\
 Mamba-2.8B-Slimpj &74.5 & 54.4 & 63.4& 44.6 &84.9 & 1381.8& 55.8& 30.6\\
\bottomrule
\end{tabular}
}
\end{table*}

\subsubsection{The Effect of Different Vision Encoders}
To evaluate the effectiveness of different vision encoders, we conduct an ablation study which is shown in Table~\ref{tab:visenc}. We study two different vision encoders, CLIP-ViT-L~\cite{Radford2021LearningTV} and SigLIP-SO~\cite{Zhai2023SigmoidLF}. The baseline models utilize Mamba-2.8B-Slimpj as LLM and vanilla MLP multimodal connectors. We can see that the CLIP-based model falls behind the SigLIP-based model in most benchmarks except the MME benchmark, where the CLIP-based model surpasses the SigLIP-based model by a large margin. Considering the comprehensive performance, we choose SigLIP-SO as the vision encoder to build the final VL-Mamba.

\begin{table*}[th!]
\caption{Ablation study of the vision encoder.
}
\label{tab:visenc}
\centering
\renewcommand{\arraystretch}{1.15}
%\scalebox{0.76}{
\resizebox{0.85\linewidth}{!}{
\begin{tabular}{l| cccc | cccc }
\toprule
Method & VQA$^\text{v2}$ & GQA & SQA$^\text{I}$ & VQA$^\text{T}$ & POPE & MME & MMB & MM-Vet \\
\midrule
 CLIP-ViT-L~\cite{Radford2021LearningTV} &74.5 & 54.4 & 63.4& 44.6 &84.9 & 1381.8& 55.8& 30.6\\
 SigLIP-SO~\cite{Zhai2023SigmoidLF} &76.7 & 55.4 & 66.3& 47.5 &85.2 & 1349.4& 56.4& 30.9\\
\bottomrule
\end{tabular}
}
% \vspace{-1mm}
\end{table*}

\subsubsection{Ablation on Different MMC Architectures}
We also explore the impact of different architectures of Multimodal Connector (MMC). We evaluate three different MMC variants: MLP, VSS-MLP, and VSS-L2.
As shown in Table~\ref{tab:arch-mmc}, by comparing the three architectures, we observe that VSS-L2 shows relatively better performance on most benchmarks, especially on the VQA$^\text{T}$, MME, MMB, and MM-Vet. The scores are 48.9, 1369.6, and 32.6 respectively, which proves the effectiveness of the VSS module combined with linear layers. Note that these models utilize SigLIP-SO as the vision encoder, Mamba-2.8B-Slimpj as the language model and Bi-directional selective scan mechanism.

\begin{table*}[th!]
\caption{Ablation study of the different architectures of MMC.
}
% \vspace{-6mm}
\label{tab:arch-mmc}
\centering
\renewcommand{\arraystretch}{1.15}
%\scalebox{0.76}{
\resizebox{0.85\linewidth}{!}{
\begin{tabular}{l| cccc | cccc }
\toprule
Method & VQA$^\text{v2}$ & GQA & SQA$^\text{I}$ & VQA$^\text{T}$ & POPE & MME & MMB & MM-Vet \\
\midrule
 MLP &76.7 & 55.4 & 66.3& 47.5 &85.2 & 1349.4& 56.4& 30.9\\
 VSS-MLP &76.7 & 54.9 & 65.4& 45.6 &85.3 & 1335.8& 56.4& 30.6\\
 VSS-L2 &76.6 & 56.2 & 65.4& 48.9 &84.4 & 1369.6& 57.0& 32.6\\
\bottomrule
\end{tabular}
}
\end{table*}

\subsubsection{Ablation on Different Scan Mechanisms}
We compare two scan mechanisms Bidirectional-Scan Mechanism (BSM) and Cross-Scan Mechanism (CSM) in the MMC module.
As shown in Table~\ref{tab:scan}, although BSM and CSM perform similarly in some benchmarks, such as they all score 76.6 in the VQA$^\text{v2}$, BSM exhibits superior performance in most benchmarks. Especially on the MMB benchmark, BSM scored 1369.6, 5.6 points higher than CSM, highlighting its strength in processing 2D vision information for multimodal learning tasks.

\begin{table*}[th!]
\caption{Ablation study of the scan mechanisms.
}
% \vspace{-6mm}
\label{tab:scan}
\centering
\renewcommand{\arraystretch}{1.15}
%\scalebox{0.76}{
\resizebox{0.99\linewidth}{!}{
\begin{tabular}{l| cccc | cccc }
\toprule
Method & VQA$^\text{v2}$ & GQA & SQA$^\text{I}$ & VQA$^\text{T}$ & POPE & MME & MMB & MM-Vet \\
\midrule
 Bidirectional-Scan Mechanism (BSM) &76.6 & 56.2 & 65.4& 48.9 &84.4 & 1369.6& 57.0& 32.6\\
 Cross-Scan Mechanism (CSM)  &76.6 & 55.8 & 64.2 & 48.8 &85.0 &1364.0& 56.3& 31.1\\
\bottomrule
\end{tabular}
}
\end{table*}

\section{Limitation}
In this paper, we are focusing on effectively applying the 2D selective scan for multi-modal connector in the VL-Mamba, without exploring the training data that would significantly affect the benchmark performance. In the future, we will study how to utilize higher-quality training data to further improve the performance of VL-Mamba.

\section{Conclusion}
In this paper, we introduce VL-Mamba, the first work that explores the state space model Mamba to solve multimodal learning tasks. The VL-Mamba consists of a language model, a vision encoder, and a multimodal connector. To be specific, we utilize the pre-trained Mamba Large Language Model (Mamba LLM) as the language model. Then, we study three architectures of MultiModal Connector (MMC) and introduce a Vision Selective Scan (VSS) module in MMC to bridge the gap between 2D non-causal image information and the inherent causal modeling capabilities of state space models (SSMs). 
In the VSS module, we propose two 2D scan mechanisms: the Bidirectional Scanning Mechanism (BSM) and Cross Scanning Mechanism (CSM).
We conduct extensive experiments on eight multimodal benchmarks and achieve comparable performance with some SoTA MLLMs, and we also conduct ablation studies to evaluate the effectiveness of language variants, different vision encoders, different MMC architectures, and different scan mechanisms. The results demonstrate the effectiveness of our proposed model and prove the potential of the SSMs applied to multimodal learning.

\bibliographystyle{abbrv}
{
	\small
	\bibliography{main}
}

\end{document}